%% file: main.tex
\documentclass[11pt,a4paper]{article}
\usepackage[hyperref]{emnlp-ijcnlp-2019}
\usepackage{times}
\usepackage{latexsym}
\usepackage{times}
\usepackage{latexsym}
\usepackage{graphicx}
\usepackage{comment}
\usepackage{color}
\usepackage{paralist}
\usepackage{xspace}
\usepackage[utf8]{inputenc}
\usepackage[bulgarian, russian, greek, english]{babel}
\usepackage{booktabs}
\usepackage{amsmath}
\usepackage{amssymb}
\usepackage{pgfplots}
\pgfplotsset{compat=newest}
\usepackage{url}
\usepackage{dblfloatfix}
\usepackage{array}

\newcommand{\Table}[1]{Table~\ref{#1}}

\newcommand{\Figure}[1]{Figure~\ref{#1}}

\newcommand{\Section}[1]{Section~\ref{#1}}

\aclfinalcopy 

\title{CodeSwitch-Reddit: Exploration of Written Multilingual Discourse \\ in Online Discussion Forums}

\author{
	Ella Rabinovich \qquad \qquad \qquad 
	Masih Sultani \qquad \qquad \qquad 
	Suzanne Stevenson 
	\vspace{0.1cm} \\
	Dept. of Computer Science, University of Toronto, Canada
	\vspace{0.075cm} \\
	\texttt{\{ella,masih,suzanne\}@cs.toronto.edu}
}

\date{}

\begin{document}
\maketitle
\begin{abstract}
In contrast to many decades of research on oral code-switching, the study of written multilingual productions has only recently enjoyed a surge of interest.  Many open questions remain regarding the sociolinguistic underpinnings of written code-switching, and progress has been limited by a lack of suitable resources.
We introduce a novel, large, and diverse dataset of written code-switched productions, curated from topical threads of multiple bilingual communities on the Reddit discussion platform, and explore questions that were mainly addressed in the context of spoken language thus far. We investigate whether findings in oral code-switching concerning content and style, as well as speaker proficiency, are carried over into written code-switching in discussion forums. The released dataset can further facilitate a range of research and practical activities.
	
\end{abstract}

\input{1-introduction.tex}
\input{3-dataset.tex}
\input{4-experiments.tex}
\input{2-relatedwork.tex}
\input{summary.tex}

\section*{Acknowledgments}
This research is supported by an NSERC Discovery Grant RGPIN-2017-06506 to S. Stevenson. We are thankful to Joel Tetreault for providing us the formal-informal GYAFC parallel corpus. We are also grateful to our anonymous reviewers for their insightful and constructive feedback.

\bibliographystyle{acl_natbib}
\bibliography{main}

\end{document}

%% file: 1-introduction.tex
\section{Introduction}
\label{introduction}
Multilingual communities adopt various communicative strategies that navigate among multiple languages. One of the most notable of such strategies is \textit{code-switching} (CS) -- when a bilingual mixes two or more languages within a discourse, or even within a single utterance, as in:

\vspace{.05in}

\noindent\selectlanguage{greek}εγώ θα το ήθελα, θα ήταν το \selectlanguage{english}guilty pleasure \selectlanguage{greek}μου\selectlanguage{english} \\
\noindent[ \textit{I would like it, it would be my} guilty pleasure ]

\vspace{.1in}
The sociolinguistic underpinnings of code-switching as an \textit{oral} conversational strategy have been investigated extensively for many decades.  The purposes of oral CS have been shown to range from practical considerations of domain-specific lexical knowledge, to indications of register and nuanced meanings
\citep[see][for a recent review]{gardner2009sociolinguistic}.
Oral code-switching is thus known to be a multifaceted component of successful communication in a multilingual community, interacting with factors of speaker proficiency, and both style and content of utterances.

By contrast, the analysis of \textit{written} code-switching has only recently enjoyed a surge of interest, and remains seriously under-studied \cite{sebba2012language}.  Written text often differs greatly from conversation in its levels of both spontaneity and formality, and findings thus far have differed in their conclusions regarding the extent to which various genres of written text reflect the same communicative functions of CS as observed in oral conversation \citep[e.g.,][]{mcclure2001oral,chan2009english, gardner2015code}.

The growing popularity of social media and online discussion platforms poses both opportunities and new research questions regarding written code-switching.  Global online forums, in which English is a lingua franca, not only draw on but create wide-reaching multilingual communities.  The resulting communications lead to a wealth of data that potentially includes a large amount of code-switching across multiple language pairs \citep[e.g.,][]{solorio2014overview, diab2016proceedings, aguilar2018proceedings}. Moreover, communication on discussion platforms often resembles a hybrid between speech and more formal writing \citep{sebba2013multilingualism}.  These differing characteristics lead to new research questions regarding the extent to which findings from oral CS carry over to these online interactions.

Research is only just beginning to grapple with these issues. Computational work on code-switching in online venues has largely focused on the practical challenges that multiple interleaved languages pose to the application of standard NLP tools \citep[as reviewed in][]{ccetinouglu2016challenges}, rather than on the communicative purposes of CS. More broadly, computational investigation of the sociolinguistic aspects of written CS is dominated by studies conducted with a limited number of language-pairs and/or authors \citep[e.g.,][]{sebba2012language}, thereby constraining the nature of questions that can be addressed with this data. 

Our work here seeks to address these gaps in the study of code-switching in online interactions. We begin by introducing the \textit{CodeSwitch-Reddit} corpus:\footnote{All data and code are available at \url{https://github.com/ellarabi/CodeSwitch-Reddit}} a novel, large, and diverse dataset of written code-switched productions, carefully curated from topical threads of multiple (including understudied) bilingual communities on the Reddit discussion platform.\footnote{\url{https://www.reddit.com/}} The main corpus comprises over $135$K code-switched messages by over $20$K unique authors, spanning five language-pairs, with average post length of $75$ tokens.

The uniform nature of our data (written communication from a single online discussion platform), as well as its ample size, pose novel opportunities for large-scale empirical investigation of research questions on code-switching -- questions that have thus far been mainly addressed in the context of oral language.  As a first study, here we explore fundamental questions about both the content and style of code-switched posts, as well as about the English proficiency level of authors who frequently code-switch. Due to the size and breadth -- yet homogeneity -- of our corpus, we can explore these questions with appropriate comparisons to monolingual posts in the same genre and register, and by the same authors.

The contribution of this work is, therefore, twofold: First, we construct a novel code-switching corpus, whose size, number of language pairs, and diversity of content (consisting of posts of unrestricted length in a wide variety of topic areas) make it a desirable testbed for a range of research questions on code-switching in online discussion forums.  Second, we demonstrate the usefulness of this dataset through an empirical investigation that sheds new light on postulated universals of code-switching -- involving linguistic proficiency, style, and content -- when inspected through the lens of online communication.

%% file: 3-dataset.tex
\section{Compilation of CodeSwitch-Reddit}
\label{dataset}

One of the contributions of this work is the CodeSwitch-Reddit corpus -- a large and diverse corpus comprising multilingual posts from online discussion forums.  We selected Reddit for construction of our corpus due to its size (over $200$M unique users), its structure of content-focused sub-communities (over $100$K active subreddits), and the unrestricted length of its posts (in contrast to Twitter, e.g.).  This enabled us to extract a large amount of code-switching in natural contexts, within multiple language pairs and by numerous authors, and with access to much monolingual text by the same authors for comparative purposes.  Because English is the lingua franca of Reddit, we restricted our focus to CS between English and another language, ensuring commonality in one language across all CS posts.

\subsection{Initial Data Extraction}

Effective use of CS relies on interactions between bilinguals, hence we needed to identify subreddits likely to contain posts in multiple languages.  We observed that country-specific subreddits (e.g., r/greece and r/philippines) often contained posts both in English and in the language of the country specified (e.g., Greek and Tagalog, respectively).  We thus restricted our extraction to all country-specific subreddits, except for countries with English as a national language, e.g., r/Australia.

We collected all posts (both initial submissions and subsequent comments/replies) from these subreddits using an API\footnote{\url{https://github.com/pushshift/api}} designed for searching Reddit content.  Among the properties collected for each post are the unique userid of the author, subreddit name, date of posting, and text of the message. Additional metadata properties include information regarding the full conversational chain of a post (e.g., links to `parent' messages), facilitating planned further extension of our corpus with contextual information.  This final raw dataset consisted of over $6.88$M posts from $71$ country-specific subreddits.

\subsection{Identifying Code-switching Posts}
We first identified posts that contained multiple languages, of which one was English, by using polyglot,\footnote{\url{https://polyglot.readthedocs.io}} a probabilistic tool that identifies the languages present in a (multilingual) text, along with their approximate percentage of the total text. This yielded a set of posts where two languages -- English and another one -- were detected. At this stage, we removed posts shorter than five tokens and those containing weblinks (which misled the language identification tool).

Second, we needed to automatically determine which of these multilingual posts actually contained code-switching.  We followed a relatively strict notion of CS \citep{barbara} as a fluid alternation between two languages in an author's own words.  Due to the inherently conversational nature of Reddit discussions, authors may include in their own post (parts of) another message they are replying to. We did not consider a post CS if the author's own text was in a single language, but the `reply-to' text contained another language. In addition, we aimed to exclude from our code-switching corpus any posts whose only alternation between two languages was a use of: named entities (e.g., `Amazon'); quotes (from websites, books, movies, etc.; e.g., ``Don't cry because it's over, smile because it happened.''); or translations (text in one language along with its equivalent in the other language).

To enforce this definition of CS, we applied the following filters to our set of multilingual posts.  Because our dataset was plentiful, we aimed at achieving high precision (occasionally at the cost of recall) to ensure that resulting posts were likely to have true examples of code-switching.

\paragraph{Replies:} A reply-to section is typically preceded by a unique sequence of characters. We used a regular expression to remove all reply-to section(s).

\paragraph{Named entities:} We removed all named entities identified by a multilingual Named-Entity Recognizer (NER) available in spaCy.\footnote{\url{https://spacy.io/}}  

\paragraph{Quotes:} Quotation marks are used both to indicate actual quotes, as well as to convey emphasis of a word or phrase, so they are a clear but noisy cue.  We used a regular expression to locate text within quotation marks, and remove it if longer than 5 tokens (since quotation marks around short phrases were often for emphasis). Other methods for quoting, such as copy and pasted text, could not be reliably identified; these are the largest source of false positives of CS in our corpus.

\paragraph{Translations:}
Here we sought to identify words that indicated that text in one language was likely a translation from the other language. Exploring lexical characteristics typical to r/translator, a subreddit for translation requests, we constructed a list of such words, and removed from our dataset all posts including any word on this set. The list of terms can be found in supplemental material (A.1).

\begin{table*}[hbt]
\centering
\resizebox{16cm}{!}{
\begin{tabular}{l}
Not valid enough for me. 7 PM pa daw kasi out niya at hihintayin ko pa rin daw siya eh \\ 
lagi ko namang ginagawa yun kapag magdidinner kami after work. So what's the difference this time? \\ \hline
Intreb pentru un prieten, care inca scrie SQL, si va scrie mult timp de acum inainte... \\
FYI, relational databases ain't gonna go away any time soon... \\ \hline
\selectlanguage{greek}Πράγματι, ήταν \selectlanguage{english} too good to be true. \\
\end{tabular}
}
\caption{Examples of code-switched posts in English-Tagalog, English-Romanian, and English-Greek.}
\label{tbl:cs-examples}
\end{table*}

\vspace{.1in} 
After obtaining a set of `clean' posts by removing sections/posts identified by these filters, we applied the polyglot tool once again.  If two languages (English and another) were still present, we identified the posts as code-switched. In order to retain as much contextual information as possible, we reinserted back into the code-switched posts any named entities and quotes that were removed by the filters above.  Table~\ref{tbl:cs-examples} presents some examples of code-switched utterances in the dataset, both intra- and inter-sentential.

\subsection{Precision of CS Identification}

The method above for identifying code-switched posts may lead to false positives, due to either polyglot errors or inadequacies in our filters.  To evaluate the precision of the resulting corpus, we obtained manual annotations, as actual CS or not, of a random sample of the posts.\footnote{We did not evaluate recall, given that our goal here was to collect a large set of CS, not to detect all CS.}

For language pairs with a large amount of data in the compiled dataset -- English plus each of Tagalog, Greek, Romanian, Indonesian, Russian, Spanish, and French -- we used FigureEight,\footnote{\url{https://www.figure-eight.com/}} a crowd sourcing platform designed to support large annotation tasks.  The tasks were split by individual language-pairs, with each including $500$ randomly sampled posts from that pair in our corpus.  Aiming at reaching bilingual speakers, the tasks were released to countries where the non-English language of the pair was a national language of that country (e.g., Greece for English-Greek).

Annotators were provided with detailed instructions and plentiful examples to indicate precisely what we considered to be code-switching; sample guidelines can be found in supplemental material (A.2).  Only those who successfully passed a (manually compiled) quiz were allowed to perform the task.  We asked the annotators to indicate yes/no whether an utterance was code-switched or not.  In the case of a `no' answer, they were prompted to provide a reason from these options: (1) no second language present; or, the presence of a second language is due to (2) a named entity, (3) a quote, or (4) other.

Each post was annotated by three annotators, and $100$\% agreement (three `yes' or three `no') was achieved on 83.4\% of the posts across all tasks. English-Indonesian and English-Tagalog had the highest percentage of unanimous annotation -- 96\% and 95\%, respectively, followed by Greek (85\%), Romanian (82\%), and Russian (60\%).\footnote{Manual inspection of English-Russian annotations revealed a wide range of (in part, subjective) usage cases employing CS; this sub-corpus accounts for $1.4$\% of our data.} A majority of `yes' answers indicated a true positive (actual CS post), and other patterns a false positive (not code-switching).  The English-Tagalog and English-Indonesian tasks yielded 99\% precision, the English-Romanian and English-Greek tasks 87\%, and the English-Russian task 85\%. The English-Spanish task had 70\% of posts identified as code-switched, and the accuracy of English-French was extremely low; on manual inspection this appeared to be due to the high extent of shared lexical items, misleading the language identification tool.

\begin{table*}[hbt]
\centering
\resizebox{16cm}{!}{
\begin{tabular}{lrrr||lrrr}
\multicolumn{1}{l}{language-pair} & \multicolumn{1}{c}{authors} & \multicolumn{1}{c}{posts} & \multicolumn{1}{c||}{avg post len} & \multicolumn{1}{l}{language-pair} & \multicolumn{1}{c}{authors} & \multicolumn{1}{c}{posts} & \multicolumn{1}{c}{avg post len} \\ \hline
English-Tagalog & 12,159 & 88,038 & 74.6 & English-Spanish & 2,461 & 4,225 & 105.4 \\
English-Greek & 3,032 & 24,585 & 59.6 & English-Turkish & 635 & 1,459 & 85.3 \\
English-Romanian & 2,516 & 10,008 & 104.5 & English-Arabic & 582 & 1,018 & 65.8 \\
English-Indonesian & 1,851 & 10,354 & 84.5 & English-Croatian & 437 & 632 & 104.6 \\
English-Russian & 934 & 2,055 & 54.3 & English-Albanian & 187 & 482 & 86.2 \\ \hline
total & 20,492 & 135,313 & 75.5 & & 4,302 & 7,816 & 89.5 \\
\end{tabular}
}
\caption{Statistics of the main CodeSwitch-Reddit corpus (left) and the additional CS dataset (right).}
\label{tbl:dataset-stats}
\end{table*}

Due to difficulty in securing bilingual workers on FigureEight, additional language pairs (with a smaller amount of data) were evaluated by an in-house team of annotators, with a relevant bilingual background where possible, or using an automatic translation tool. Five language pairs were identified as comprising between 70\% and 85\% code-switched posts.

\subsection{Final Corpus and Additional Datasets}
\label{final-datasets}

The final CodeSwitch-Reddit corpus includes the five language pairs that had $85$\% or higher precision.  Table~\ref{tbl:dataset-stats} (left) reports details on the corpus, which has over $135$K posts, with an average of $75$ tokens per post, thereby introducing a unique resource facilitating further research in this field. (A sample of English-Tagalog posts are in supplemental materials.) Table~\ref{tbl:dataset-stats} (right) reports the details of the five additional language pairs that had at least $70$\% precision.  Because of the lower quality, we omit these language-pairs from our experiments below.  However, we recognize the potential usefulness of this additional data, and release it as an addendum to our main corpus, for possible further cleanup and preprocessing.

We also compiled an additional dataset of English \textit{monolingual} posts from the same country-specific subreddits as our code-switched corpus.  Here, as above, we excluded posts with weblinks and posts shorter than five words.  This dataset serves as an appropriate comparison set for the code-switching corpus, and is used for some analyses in Section~\ref{sec:experiments}.

%% file: 4-experiments.tex
\section{Sociolinguistic Aspects of Online CS}
\label{sec:experiments}



A large body of research has highlighted code-switching as a strategy used to shape the dynamics of the sociolinguistic situation, including aspects of identity, interpersonal relationships, and formality \citep[e.g.,][]{gardner2009sociolinguistic}.  In addition, studies have differed in their conclusions on whether those who code-switch in speech are highly adept bilinguals \citep[e.g.,][]{kootstra2012priming}, or use code-switching to address lexical deficiencies \citep[e.g.,][]{poulisse1989use}.  Using our new corpus, we aim to explore how such findings from the research on CS in oral communication apply in online written communication. Namely, we test whether code-switched text on the Reddit discussion platform is typified by unique topical and informality markers, and whether authors who frequently mix languages in their posts differ in their English proficiency level from those who do not. 
Our dataset, comprising communication in multiple language-pairs, as well as monolingual English writing by thousands of authors within the same register introduce an appropriate testbed for addressing these questions.

\subsection{Topical Preferences in CS Text}
\label{sec:topics}

We take a topic modeling approach to explore whether CS productions are characterized by a unique topical flavor, when compared to monolingual messages.  Specifically, we compare two topic models, one created using English text from the CodeSwitch-Reddit corpus, and a second using the monolingual dataset from the same subreddits (see \Section{final-datasets}).  We use only the text of each dataset from \textit{a common set of authors} whose posts appear in both, thereby ensuring that any differences we find are likely due to the context of code-switching rather than individual differences in content between authors.

We identified a set of $4,843$ authors who posted both code-switched and monolingual messages in our datasets, resulting in $58,568$ CS posts and $58,603$ monolingual posts, having a minimum of $50$ tokens and averaging $108.7$ and $103.7$ tokens per post, respectively. We tokenized and lemmatized texts, excluded English stopwords, and removed name entities and tokens other than nouns, verbs, adjectives and adverbs using python spaCy tools. We applied a dictionary-based approach in order to filter out non-English words; we removed all words ranked higher than $10,000$ in the word-rank list built from the English Wikipedia corpus.\footnote{\url{https://dumps.wikimedia.org/enwiki}}

\begin{table*}[hbt]
\centering
\resizebox{16cm}{!}{
\begin{tabular}{
>{\centering\arraybackslash}p{1.5cm}
>{\centering\arraybackslash}p{1.5cm}
>{\centering\arraybackslash}p{1.5cm}
>{\centering\arraybackslash}p{1.5cm}
>{\centering\arraybackslash}p{1.5cm}|
>{\centering\arraybackslash}p{1.5cm}
>{\centering\arraybackslash}p{1.5cm}
>{\centering\arraybackslash}p{1.5cm}
>{\centering\arraybackslash}p{1.5cm}
>{\centering\arraybackslash}p{1.5cm} }
\multicolumn{5}{c|}{Topics with highest coherence scores in code-switched posts} & \multicolumn{5}{c}{Topics with highest coherence scores in monolingual posts} \\  \hline
CS-1 & CS-2 & CS-3 & CS-4 & CS-5 & M-1 & M-2 & M-3 & M-4 & M-5 \\
student & friend & make & give & phone & thing & vote & problem & learn &  plan \\
exam & feel & thing & parent & check & person & power & change & student & price \\
study & girl & happen & live & driver & love & political & issue & study &  cost \\
teacher & love & talk & chance & internet & call & support & reason & experience & cheap \\
grade & date & wrong & reason & price & wrong & rule & fact & teach &  sell \\
graduate & close & word & plan & mall & happen & majority & deal & education & stay \\
pass & break & hate & late & shop & hate & politician & current & skill & travel \\
subject & relationship & joke & decide & traffic & understand & citizen & control & teacher & visit \\
program & happy & sense & child & store & kind & election & matter & research & expensive \\
review & meet & mind & baby & brand & accept & leader & point & test & compare \\
\end{tabular}
}
\caption{Most coherent topics identified in code-switched and monolingual posts by the same set of authors.}
\label{tbl:topic-modeling}
\end{table*}

\begin{table*}[!b]
\centering
\begin{tabular}{l}
Entahlah, I'm self diagnosing, but maybe I'm on the spiral to depression... \\
$[$ \textit{I've no idea}, I'm self diagnosing, but maybe I'm on the spiral to depression... $]$ \\ \hline
If you somehow read this... \selectlanguage{russian}Лена, я дурак. Мне очень жаль. Я очень по тебе скучаю. \\
$[$ If you somehow read this... \textit{Lena, I'm a fool. I'm sorry. I miss you so much.} $]$ \\ \hline
Lovely.... whataboutism at it's finest. \selectlanguage{greek}Επισης κάνεις κάτι ατυχέστατες υποθέσεις. \\
$[$ Lovely.... whataboutism at it's finest. \textit{You're also doing something wrong.} $]$
\end{tabular}
\vspace{-0.05in}
\caption{Examples of CS posts (with \textit{translations}) associated with emotion, sentiment, and relationships.}
\label{tbl:relationships}
\end{table*}

We used a publicly-available topic modeling tool \citep[MALLET,][]{McCallumMALLET} to identify the prevalent discussion topics in the CS and monolingual messages. Each topic is represented by a probability distribution over the entire vocabulary, where terms more characteristic of a topic are assigned a higher probability. A common way to evaluate a topic learned from a set of documents is by computing its \textit{coherence score} -- a statistical measure reflecting the quality of a topic \cite{newman2010automatic}. The quality of a learned model is then estimated by averaging the scores of its individual topics -- the \textit{model} coherence score.  We selected the optimal number of topics for each set of posts by maximizing its model coherence score, resulting in $17$ topics for CS posts (score of $0.58$), and $21$ topics for monolingual posts (score of $0.60$). 

\paragraph{Qualitative analysis:} We first examine similarities and differences across the two topical distributions by extracting the top $5$ topics -- those with the highest individual coherence scores -- in each of the code-switched and monolingual models. \Table{tbl:topic-modeling} presents the $15$ most probable words for these top-five topics in each of the code-switched and monolingual models (on the left and right sides, respectively); topics within each are ordered by decreasing coherence score (left to right). 

While both CS and monolingual posts include a topic expressing \textit{sentiment about events} (topics CS-3 and M-1), CS posts have two further topics that particularly discuss \textit{relationships} and \textit{family} (CS-2, CS-4).  This suggests that the choice to switch languages on Reddit can delineate social, personal, and emotional territory.  Other topics typify both CS and monolingual posts, discussing \textit{studies} (CS-1, M-4) and \textit{shopping/travel} (CS-5, M-5).  Discussions on \textit{politics/issues} are prevalent only in monolingual communication (M-2, M-3), again indicating that CS posts are more focused on personal experiences.

Table \ref{tbl:relationships} presents a few examples of code-switched posts associated most with the \textit{relationship} topic in the model.  It is worth noting that this is one of the subjects most prominently associated with code-switching in speech. The fluid and seamless interleaving of languages highlights the way that emotion, sentiment, and personal experience can guide lexical choice in written communication on the Reddit platform.

\paragraph{Quantitative analysis:}
We further assess the extent of topical distinctions between code-switched and monolingual posts by performing statistical significance analysis. Namely, we randomly partition the monolingual set of posts into two equal-sized subsets and test whether the two parts exhibit higher mutual similarity than (on average) with their code-switched counterpart.

Formally, given two topic models $M_1$ with topics $T_1^1, ..., T_1^N$, and $M_2$ with topics $T_2^1, ..., T_2^K$, where $N$ and $K$ are the number of topics in $M_1$ and $M_2$, respectively, we define the similarity between two models as the average of their pairwise topic similarities:

\vspace{-0.2in}
\begin{equation}
sim(M_1,M_2) = \frac {1}{N{\times}K}\sum JSC(T_1^i, T_2^j)
\label{eq:sim-avg}
\end{equation}

\noindent 
where $i\in[1..N], j\in[1..K]$, and the Jaccard similarity coefficient (JSC) is used to measure the extent of overlap between the top $100$ terms associated with each of $T_1^i$ and $T_2^j$.  

Given model $M_{cs}$ learned from the code-switched corpus, and models $M_{en}^1$ and $M_{en}^2$ from the two partitions of the monolingual posts, we can then compute two similarity metrics:

\vspace{-0.1in}
\begin{equation}
avg(sim(M_{cs}, M_{en}^1),sim(M_{cs}, M_{en}^2))
\label{eq:sim-inter}
\end{equation}
and
\vspace{-0.05in}
\begin{equation}
sim(M_{en}^1, M_{en}^2)
\label{eq:sim-intra}
\end{equation}

We repeated the experiment with $30$ random partitions of the monolingual corpus, resulting in mean \textit{inter}-corpora similarity (Equation~\ref{eq:sim-inter}) of $0.030$, and mean \textit{intra}-corpus similarity (Equation~\ref{eq:sim-intra}) of $0.037$.\footnote{The extremely low similarity scores reflect the (desirable) low average similarity between truly different topics.} A Wilcoxon rank-sum test for significance of the differences yielded a $p$-value of $2.87e\mbox{-}11$. Performing the same analysis with \texttt{max} instead of \texttt{average} topical similarity in Equation~\ref{eq:sim-avg} (i.e., estimating sub-corpora topical similarity through two maximally similar topics), yields mean \textit{inter}-corpora similarity of $0.384$, and mean \textit{intra}-corpus similarity of $0.470$, with $p$-value of $4.99e\mbox{-}7$. We, therefore, conclude that significant differences can be observed in topical preferences in code-switched messages compared to monolingual texts in our datasets.

\subsection{(In)formality of code-switched content}

Oral code-switching has been shown to be a marker of informality \cite{genesee1982social, myers1995social}. While user-generated content on the Reddit discussion platform is by and large considered informal, in this section we test whether (presumably subtle) formality differences can be observed in CS posts compared to monolingual ones. In accord with findings in spoken language, we expect more informal context to stimulate more frequent language mixing.

We use the formal-informal GYAFC parallel dataset by \citet{rao2018dear} to extract a set of markers indicative of informal style. Originally collected from the Yahoo Answers platform, the corpus comprises over $50$K sentence-pairs in the domain of Entertainment\&Music and Family\&Relationships, where each original \textit{informal} sentence is paired with a manually generated \textit{formal} counterpart. The parallel nature of the corpus facilitates extraction of a clean set of stylistic markers characteristic of informal text, while abstracting away from content. We use the
\textit{log-odds ratio informative Dirichlet prior} \citep{monroe2008fightin} to discover markers appearing more in informal style, by comparing the informal part of the GYAFC dataset to its formal part. We used the strict log-odds score of $-5$ as a threshold for collecting terms associated with informal writing. Among the markers identified by this procedure were \{\textit{!, ..., u, dont, n't, its, just, ur, im, thats, like, na, \&, really, got, lol, cant, ya, coz, alot, but, yea, doesnt, so, kinda, hey, dude, 've, pretty}\}.

Using the same sets of posts as in \Section{sec:topics}, we tested our hypothesis by inspecting these markers of informality in CS posts compared to monolingual posts by the same set of authors. Specifically, each author is assigned two measurements, reflecting the frequency of the informality markers in the entire collection of their code-switched posts and of their monolingual English posts. Frequency was computed by normalizing the total count of informality markers in an author's text (concatenation of individual posts) by its length.

\begin{figure}[hbt]
\centering
\includegraphics[scale=0.35]{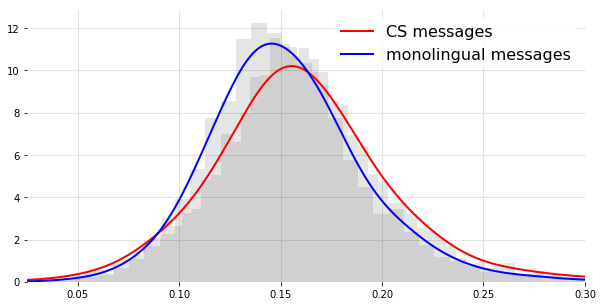}
\vspace{-0.05in}
\caption{Kernel density estimate of informality markers' frequency in CS and monolingual messages.}
\label{fig:formality}
\end{figure}

The mean frequency of informality markers in CS posts was $0.160$, while in monolingual posts by the same authors was $0.153$. Figure \ref{fig:formality} illustrates the kernel density estimation of these frequencies: the subtle deviation in formal style is mirrored by the slight right shift of the density function for code-switched texts.  Moreover, a (paired difference) Wilcoxon signed-rank test for statistical significance of the difference in CS and monolingual posts resulted in a within-subjects effect size of $0.15$ and a very low $p$-value ($1.01e\mbox{-}23$).
Despite the generally high level of informality of Reddit postings, we see a small but reliably detected difference of even greater informality in CS utterances, mirroring the observations in oral CS being associated with informal speech.  

\subsection{Code-switching and language competence}

Theoretical and experimental studies have differed in considering code-switching as a strategy of extremely proficient bilinguals, or as a coping mechanism of those with deficiencies in one of the languages (see Section~\ref{sec:related-work}).
We address this issue in the context of written communication by computing a range of lexical and grammatical metrics from utterances of authors who tend to mix languages frequently, compared to those who do not.


We separated authors who were active contributors to bilingual subreddits into those who frequently code-switch vs.\ those who do not. We identified $288$ authors as `high code-switchers' -- those who have over $100$ messages in a country-specific subreddit, where at least $20\%$ of these posts were code-switched.  We further identified $262$ authors as `low code-switchers', having over $100$ posts but mixing languages in less than $2\%$ of their messages. Aiming at a reliable assessment of these authors' English linguistic skills, we gathered their entire monolingual English digital footprint from the Reddit platform, including their posts in all but country-specific subreddits. The collected set of messages contains an average of $1,060$ and $856$ posts per author, for high and low code-switchers, respectively.

We estimate linguistic competence by computing a set of lexical and grammatical measures commonly used for language proficiency assessment \cite{kyle2015automatically, lu2015syntactic, rabinovich2018native}. All posts were tokenized and lemmatized, and non-alphabetic tokens were excluded from the computation. We produce the following metrics:

\begin{figure*}[h!]
\centering
\includegraphics[scale=0.6]{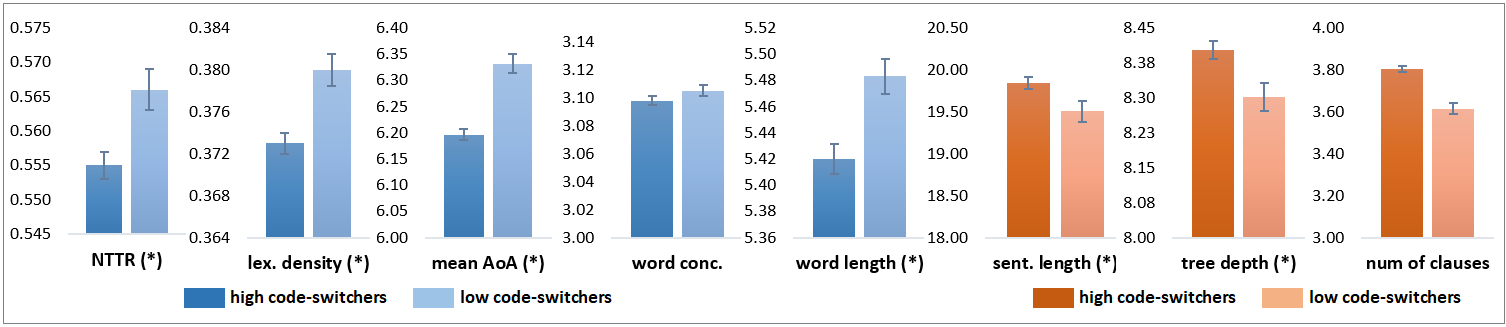}
\caption{\small{Linguistic proficiency metrics (mean and standard error; lexical on the left, grammatical on the right) of high and low code-switchers. `*' indicates significant difference in a Wilcoxon rank-sum test at the level of $p<.05$.}}
\label{fig:proficiency}
\end{figure*}

\vspace{0.1cm}
\noindent\textbf{Normalized type-to-token ratio} (NTTR) -- the ratio of the number of unique tokens and the total number of tokens, averaged over non-overlapping sliding windows of 1000 tokens.

\vspace{0.1cm}
\noindent\textbf{Lexical density} (lex.\ density) -- the ratio of the number of content tokens (excluding function words) and the number of all tokens, using $426$ function words from \citet{koppel2011translationese}.

\vspace{0.1cm}
\noindent\textbf{Mean average age of acquisition} (mean AoA) -- mean of the average AoA of tokens for over $30$K English words \citep{kuperman2012age}.

\vspace{0.1cm}
\noindent\textbf{Mean word concreteness} (word conc.) -- mean word concreteness of tokens, taken from ratings for $40$K English words \cite{brysbaert2014concreteness}.

\vspace{0.1cm}
\noindent\textbf{Mean word length} (word length) -- the average length in characters of a token.

\vspace{0.1cm}
\noindent\textbf{Mean sentence length} (sent. length) -- the average length of a sentence (in tokens).

\vspace{0.1cm}
\noindent\textbf{Mean parse tree depth} (tree depth) -- the average depth of a parse tree of a sentence, using the constituency parser of \citet{kitaev2018multilingual}.

\vspace{0.1cm}
\noindent\textbf{Mean number of clauses} (num of clauses) -- the average number of clauses in a sentence.

\Figure{fig:proficiency} presents these metrics for high and low code-switchers. We find that high code-switchers exhibit statistically significant \textit{lower} proficiency in the lexical metrics (with the exception of word concreteness level), but statistically significant \textit{higher} proficiency for two of the grammatical metrics.  Although the differences in both cases are small, they are reliably detected given the size of the corpora.  Interestingly, while the lexical measures indicate less `sophisticated' English words used by the high code-switchers, the distinctions in their grammatical complexity (reflected in longer sentences and greater parse tree depths) imply that mixing language in written communication may not necessarily be a manifestation of \textit{overall} lower linguistic competence. These intriguing findings suggest that code-witching in online forums is a multifaceted phenomenon: while bridging lexical deficiencies, it may require advanced grammatical capabilities in order to construct mixed sentences without distorting the `grammaticality' of the target utterance. These results both demonstrate the value of studying code-switching at scale, and leave room for much further investigation. Our future plans include breaking down the data by individual native languages of the authors, in order to capture differences in proficiency stemming from various L1 backgrounds.

%% file: 2-relatedwork.tex
\section{Related Work}
\label{sec:related-work}

Many studies have elucidated the communicative functions of oral code-switching.  These include conveying nuanced attitudes through linguistic choices reflecting emotion and sentiment \citep{dewaele2013multilinguals}, as well as establishing a level of (in)formality \citep{fishman1970sociolinguistics, genesee1982social, myers1995social}. Other studies have focused more on the characteristics of code-switchers themselves:  While CS has mostly been viewed as a deliberate choice of a proficient bilingual speaker, requiring adept control of two simultaneously-activated linguistic systems \citep{costa2004lexical, kootstra2012priming}, mixing two languages may also serve as a strategy for bridging lexical inefficacy in a second language \citep{grosjean1982life, faerch1983strategies, poulisse1989use}.

As noted earlier, written CS is much less studied and the extent to which these findings hold across various written genres is not clear \cite{sebba2012language}. Recent research from the sociolinguistic perspective has considered CS in an array of online genres: e.g., English-Jamaican Creole in email \citep{hinrichs2006codeswitching}, 
Malay-English language in online discussion forums \citep{mclellan2005malay}, English-Spanish bilingual blogging \citep{montes2007blogging}, and English-Chinese 
instant messaging \citep{lam2009multiliteracies}.  However, these studies have analyzed productions of a small number of authors (even a single person), and each is restricted to a single language-pair.  

Most large-scale computational work on CS in social media has focused on essential prerequisites for various NLP tasks, such as POS tagging \citep{solorio2008learning, soto2018role}, token-level language identification \citep{soto2018role, rijhwani2017estimating}, NER \citep{aguilar2018named}, language modeling \citep{chandu2018language}, automatic speech recognition \citep{yilmaz2018building}, etc.
Almost no work has addressed theoretical and sociolinguistic aspects of written multilingual discourse computationally, at scale \citep[for an exception, see][]{rudra2016understanding}.

Our work here addresses various gaps noted in the research above.  We create (and make available) a novel dataset that enables investigation of research questions on written code-switching at scale, across multiple language pairs and a large number of speakers. Moreover, we use this dataset for computational, large-scale investigation of subtle issues of linguistic content and style, as well as speaker proficiency, that have been extensively studied in the context of spoken language, and enjoyed very limited attention in written communication.

%% file: summary.tex
\section{Conclusion}
\label{conclusion}
Despite the presumed differences between code-switching in spoken and written communication, 
multilingual written discourse is understudied, partially due to the lack of adequate resources. In this work we introduced the CodeSwitch-Reddit corpus -- a large, diverse and carefully curated dataset of code-switched posts collected from multiple multilingual discussions forums on Reddit. Computational investigation of this dataset reveals topical and stylistic distinctions between code-switched and monolingual communication, as well as differences in the proficiency level of authors who mix languages frequently, compared to those who do not. Our future plans include extending this dataset with conversational context and investigating additional aspects of online written multilingual discourse in richer contexts.  

The released dataset suggests a wealth of both intra- and inter-sentential code-switched utterances, and investigation of sociolinguistic differences between the two introduces an interesting research question. As an example, while mixing languages within a sentence could be indicative of lexical deficiency, that may not necessarily be the case with inter-sentential code-switching.